\documentclass[letterpaper, 10 pt, conference]{ieeeconf}

\IEEEoverridecommandlockouts
\usepackage{graphicx} 
\usepackage{listings}
\usepackage{flushend}

\title{\LARGE \bf
Improving the lifecycle of robotics components using Domain-Specific Languages}

\author{ \parbox{4.5 in}{\centering A. Romero-Garc\'es, L.J. Manso, Marco A. Guti\'errez, R. Cintas, P. Bustos * \\
         \thanks{*This work has been partially supported by project TSI-020301-2009-2 funded by the Spanish Government and Feder funds, by the Junta de Extremadura projects IB10062, PRI09A037, PDT09A59 and PDT09A044} RoboLab, Computer and Communication Technology Deptartment, University of Extremadura, C\'aceres, Spain. \\ {\tt\small pbustos@unex.es}}
}
\begin{document}

\maketitle
\thispagestyle{empty}
\pagestyle{empty}

\begin{abstract}
There is currently a large amount of robotics software using the component-oriented programming paradigm. However, the rapid growth in number and complexity of components may compromise the scalability and the whole lifecycle of robotics software systems. Model-Driven Engineering can be used to mitigate these problems. This paper describes how using Domain-Specific Languages to generate and describe critical parts of robotic systems helps developers to perform component managerial tasks such as component creation, modification, monitoring and deployment. Four different DSLs are proposed in this paper: {\it i)} CDSL for specifying the structure of the components, {\it  ii)} IDSL for the description of their interfaces, {\it iii)} DDSL for describing the deployment process of component networks and {\it iv)} PDSL to define and configure component parameters. Their benefits have been demonstrated after their implementation in RoboComp, a general-purpose and component-based robotics framework. Examples of the usage of these DSLs are shown along with experiments that demonstrate the benefits they bring to the lifecycle of the components.

\end{abstract}

\section{INTRODUCTION}
Much effort has been placed in developing tools that provide better software reusability and scalability. When it comes to robotics software, the need for these tools is even higher due to the multidisciplinary and heterogeneous nature of the programs that are built. Component-oriented programming, which to some point can be seen as an extension of object-oriented programming, is a promising direction towards achieving these goals. Components are software modules -usually standalone programs- providing an interface for interaction with other components. This structure makes them a more autonomous and reusable concept than classes. However, components have a lifecycle that can become quite complex, especially in robotics environments~\cite{brugaliI}\cite{brugaliII}. The lifecycle of a component includes all common activities that are present during the lifecycle of any program: requirements analysis, design, implementation, unit testing, system integration, verification and validation, operation support, maintenance and disposal. However, due to the ever-changing requirements typical of robotics software, the lifecycle of robotics components is extremely active. To help developers in these tasks, specific tools are required.
\par
Model-Driven Engineering has proven to mitigate this situation~\cite{MDE}. In particular, it is worth noting the OMGs Model-Driven Architecture (MDA)~\cite{MDA}. This methodology keeps the system specification (model) separated from the system implementation. MDA models are structured by layers with different levels of abstraction. With this structure, we can build platform-independent models (PIM) which provide high-level designs, and platform-specific models (PSM) which contain those elements that depend on the final system implementation. Transformations from PIM to PSM are described in MDA so developers can obtain their low-level designs from high-level ones. Languages in MDA are called meta-models and are described using a common root meta-metamodel: the Meta Object Facility (MOF)~\cite{MOF}. The main advantage of MOF is that, once metamodels have been created, developers can benefit from model to model transformations (M2M) or model to text transformations (M2T) to obtain source code in an automatic way.
\par
In this paper we describe the introduction of MDAs concepts in the design of the RoboComp robotics framework ~\cite{robocomp}. RoboComp is a component-oriented framework built around a component model, a communications middleware, a repository structure and a set of tools used for developing and deploying components. After six years of steady development, the repository holds more than one hundred components, covering functionalities of different robotics and artificial vision topics. However,  the original design of the framework, which is not MDA-oriented,  presented several problems related to the life-cycle of the components. Our group has identified and tracked those issues of the components life-cycle that where responsible for most of non-productive developing time. These issues can be grouped in two classes: those related to code development, and those related to deployment.  Main coding issues  are the creation of new components, addition and deletion of proxies to existing components and modifications in the interface definitions. Deployment issues are mainly due to wrong values in the configuration parameters of the components and unresolved dependencies among them. In Section \ref{robocomp} more details are presented that justify the need for redesigning specific parts of the RoboComp architecture.

At the end, all these issues come from the need to manually rewrite parts of the components that could be automatically modified. This process is readily prone to errors if done by humans. The solution adopted was to redesign the component model, dividing it in two different parts: a generic one generated automatically from DSLs,  and a specific one written by the user. Four domain-specific languages (DSLs) have been created as part of the new design:  first, a Component Description Specific Language -CDSL-  for describing the main characteristics of the components; second, an Interface Description Specific Language -IDSL- to specify component interfaces independently of the underlying communications middleware; third, a Deployment Description Specific Language -DDSL- to describe the deployment process of robotic components networks; and finally, a Parameter Description Specific Language -PDSL- to describe the configuration parameters and their values, that determine the run-time behavior of the components.
\par
The rest of the paper is organized as follows. Section~\ref{related} provides an overview of the related works. Section~\ref{robocomp} introduces the RoboComp framework and why it was found necessary to incorporate DSL technologies to it. Next, the RoboComp Domain-Specific Languages are described in section~\ref{dsls}. Finally, the case study and the conclusions are detailed in sections~\ref{study} and~\ref{conclusions}, respectively.

\section{Related work}
\label{related}
Software development for robots can benefit from the use of \mbox{MDA-based} tools. The CoSMIC visual toolkit~\cite{COSMIC} is an interesting example of this. It is a complete open source MDA tool that allows the visual design, deployment and configuration of components based on the CORBA Component Model (CCM)~\cite{CCM}. However, CCM and its associated specifications are not widely used. In spite of that, the OMG aims at their future adoption within the robotics community by standardizing its novel Robotic Technology Component Specification (RTC)~\cite{RTC}, which focuses on the structural and behavioral features required by robotics software as a supplement to a general component model. In this vein, it is worth noting  \mbox{OpenRTM-aist}~\cite{OpenRTM}, a free RTC implementation that has appeared recently. \mbox{OpenRTM-aist} is a framework for robotics that provides developers with a set of tools to create and manage components. Two of these tools are Eclipse-based GUI tools: RTBuilder and RTSystemEditor. RTBuilder allows developers to create components defining their names, connectors (ports), parameters, programming language or their operating system. This tool can then be used to automatically generate source code templates. RTSystemEditor provides a mechanism to edit and configure components which are registered on a known name service. RTSystemEditor can start, stop or reset components, add and remove links and use introspection capabilities to monitor components at run time.
\par
Another interesting tool is the 3-View Component Meta-Model (V3CMM)~\cite{V3CMM}. It is a platform-independent modeling language for component-based applications that makes use of MOF-based metamodels. V3CMM provides three views that are loosely coupled and allow users to design a complete system. With V3CMM it is possible to model the static structure of components (named as structural view), the behavior of these components (coordination view) and to model the functionality of these components, described as algorithms (algorithmic view). Two of these views are based on UML~\cite{UML}: 1) the coordination view uses a simplified version based on UML state machines in order to model the different states of a component and 2) the algorithmic view, based on UML activity diagrams, executes a specific behavior depending on the current state of the component. The structural view is used to define components and their dependencies by specifying both its required and provided interfaces. Once users model the system using these three views, it is possible to perform M2M to reduce the level of abstraction and M2T transformations to automatically obtain the final source code.
\par
The SmartSoft~\cite{SMARTSOFT} robotics framework also provides an MDA tool based on a UML profile implementation. The development process starts modeling an idea at a high-level of abstraction. This model is then refined through several transformations to obtain the software components source code. First, developers have to describe the system in a model independent platform (PIM) where information about middleware, operating system, programming languages and other properties are unknown. Then the PIM is transformed to a platform-specific model (PSM), where details about middleware or operating systems are specified. This PIM is also transformed to a platform-specific implementation (PSI)  where developers can add their code and libraries. The next step is to deploy the components. In this vein SmartSoft uses the platform description model (PDM) to define the target platform properties. The model is extended with this platform information and finally the system can be run following the specified deployment model. During the development process users can guide the transformations in order to obtain a specific component by selecting the desired real-time and QoS properties of the component and the communication middleware it will use. A set of well-defined communication patterns provides the necessary abstraction from the final communication model and its reference implementation. Currently, it supports the ACE~\cite{ACE} (SmartSoft/ACE) and ACE/TAO~\cite{TAO} (SmartSoft/CORBA) communication middlewares and provides other interesting features, such as a mechanism to guarantee real-time properties using an external scheduler analyzer or dynamic wiring for components.
\par
For many years, visual modeling tools (such as UML) have been used to specify, design and document complete systems. These visual paradigms are intuitive for users (analysts, designers, etc) and are easier to understand for clients than textual models. Despite these advantages, expert developers on a particular domain sometimes feel more comfortable with textual representations, that allows them to build their models as if they were working with common programming languages. OpenRTM, V3CMM and SmartSoft also provide graphical tools to create and manage visual models but these tools do not have the ability to use textual ones. Thus, we decided to create the DSLs as textual models because they can be automatically exchanged with visual ones (i.e., developers can choose at any moment how to develop their systems, using textual or visual representations) and because its edition can also be assisted by the DSL editor (i.e., pointing to syntax and semantic errors).

\section{The need of DSLs in RoboComp}
\label{robocomp}
RoboComp is a distributed, component-oriented, tool-enhanced, robotics framework in development since 2005~\cite{robocomp}. RoboComp currently holds dozens of components spanning a hardware abstraction layer, basic navigation skills, SLAM, visual processing algorithms, planning, manipulator control, 3D object recognition and many others. It is being used by several research groups and companies. As mentioned before, components are processes  with structured interfaces~\cite{brugaliI,brugaliII}. Each component encapsulates a specific functionality and can communicate with others, giving rise to complex behaviors sustained by their dynamic interactions. 
\par
Like other robotics frameworks, RoboComp provides a wide set of tools aimed at reducing the required effort to perform everyday tasks, such as component creation, modification and deployment. When dealing with large graphs of processes containing more than a few components (e.g., when controlling mobile manipulators endowed with expressive heads), the development team needs well designed tools targeted at reducing the time wasted in common and repetitive errors. By these errors we mean those made when inserting or changing code not directly related to the behavior of the component. To design useful tools and to improve the underlying architecture, the first step is to identify error-prone tasks that could be automated or assisted. We have identified five tasks as the most error-prone and time-consuming ones:
\begin{enumerate}
\item Creation of new components.
\item Adding or removing proxies to existing components.
\item Adding or changing existing data types or methods in interface declarations.
\item Errors in configuration parameters that determine the run-time behavior of the components.
\item Unresolved dependencies when deploying graphs of connected components.
\end{enumerate}
\par 
Although there were some tools and scripts written to mitigate these issues, they lacked of the necessary generality to adapt to the continuous evolution of the component model. For instance, at some point, the model was modified so new components implemented a generic introspection interface. This interface is served by an internal thread that, at start, reads and checks the configuration parameters and, afterwards, parsimoniously monitors the execution of the working thread. These changes in the model are difficult to incorporate in the component creation scripts if they are designed as simple template transformation processes. Moreover, once a component is created, introducing changes is even more difficult, specially when several versions of the components coexist in the system.
Despite middleware-independency is underway~\cite{simpar}, RoboComp currently uses Ice~\cite{ZeroC_Ice}  as its main communication middleware. Ice is an extremely robust RPC-oriented technology that supports a rich variety of communication resources, including a push/pull data-oriented mechanism. Our goal is to evolve RoboComp towards a middleware-independent framework, with several reference implementations. The first step in this direction is the definition of an IDL that can act as a bridge between the IDLs of the final middlewares and RoboComp. This new IDL provides a syntax for construction of data types and procedures that can be safely used inside RoboComp code without relying on external dependencies of third party providers.
\par
Finally, the last source of repetitive errors is related to the deployment of components. The two main reasons are wrong or inadequate configuration values and unresolved dependencies found when deploying connected components. Most of the first type of errors can be solved if configuration parameters are described in a more structured way, including range values definition and checking. Moreover, the planning of a correct and safe deployment can be simplified if a structured description language is available, along with the necessary tools to orderly execute the involved processes.
\par
All these issues can be easily tackled by separating the code that can be potentially generated automatically from the code produced by developers. This design choice makes possible to modify the generic properties of the components without interfering with what was manually changed since the creation of a component. Moreover, this isolation would facilitate the inclusion of new features that were previously not taken into account, such as the optional use of graphical interfaces, internal state-machines, auxiliary classes or third-party libraries.
\par

\section{IMPROVING ROBOCOMP WITH DSLs}
\label{dsls}
We propose an MDA-based approach to mitigate the problems presented in the previous section providing a higher level of abstraction. In particular, our approach proposes four DSLs:  CDSL, IDSL, PDSL and DDSL. These DSLs enable users to work with RoboComp components in an intuitive way, improving the management of their lifecycle.  
\par
The tools developed for this purpose are based on Eclipse, which provides a powerful framework for developing MDA-based tools. In this vein, we have used the following Eclipse MDA based plugins: {\bf EMF}~\cite{EMF} (Eclipse Modeling Framework) which is a basic MOF implementation; {\bf Xtext}~\cite{Xtext} which is a framework to create textual representations and notations from visual models and metamodels; and {\bf MOFScript}~\cite{MOFScript}, which provides a template language to perform M2T transformations. It was decided to work with textual model representations because it allows developers to build their DSLs as they usually do when working with any other programming languages, textually, and switch to visual models when necessary. Xtext is a recent tool that facilities the creation of new DSLs and provides some interesting and useful characteristics such as code completion, syntax error checking and syntax highlighting among others. Moreover, its integration with EMF allows Xtext models to be represented as Graphical Modeling Framework~\cite{GMF} visual models at any moment.
\par
These DSLs help developers to quickly understand the structure of a component, design it or even modify it at any time in its lifecycle. Along with the languages, some tools to take advantage of them have also been developed. The RoboComp DSL Editor provides users with an Eclipse-based tool to create and manage the proposed DSLs and integrates MOFScript in order to generate code automatically. The component manager, named RCControlManager, deploys the necessary components using DDSL models.
\par
Section~\ref{study} will provide a case study covering all of the proposed DSLs and an experiment carried out to quantify the benefits of this technology in the developing time of RoboComp components.
\subsection{CDSL}
\label{cdsl}
RoboComp provides both client/server and publish/subscribe communication models. In order to establish communication, components must perform different operations. For example, if a component is required to perform remote calls to other components, its code needs to: {\bf a)} include the definition of the proxy classes corresponding to the interfaces it is going to connect to; {\bf b)} read from the configuration file how to reach the remote component; {\bf c)} create the proxy object using the previously read configuration; {\bf d)} provide the proxy object to the classes that will be using it. Similar scenarios exist when providing new interfaces, subscribing or publishing new topics (see~\cite{robocomp} for more details). In RoboComp, this code is automatically generated by a Python script when the component is created for the first time. However, until the adoption of MDA-based techniques, if the requirements of a component changed after its creation (a considerably common scenario), it had to be manually done.
\par
All these situations made it difficult to maintain several components because their source code depends on these parameters. To solve these problems, we have developed a DSL to create and modify component properties. This DSL is called Component Description Specific Language (CDSL) and allows users to create and maintain their component descriptions from a textual model. CDSL files contain information about communication parameters such as proxies, interfaces and topics used by the components, their dependencies with external classes and libraries, the optional support of Qt graphical interfaces, the programming language of the component, and an optional SCXML file path for embedding a state machine in the component.
\par
\begin{figure}[t!]
\center
\includegraphics[width=0.49\textwidth]{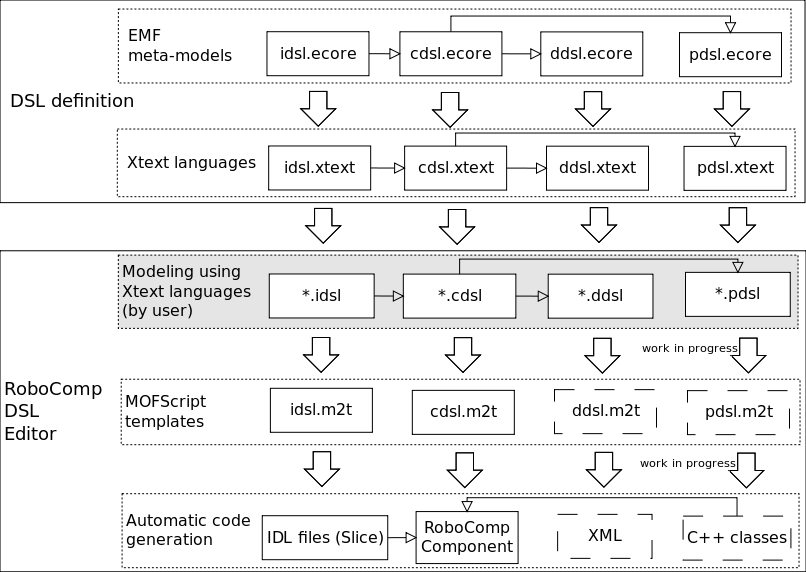}
\caption{Development process of CDSL, DDSL, IDSL and PDSL}
\label{dsl-development}
\end{figure}

Figure~\ref{dsl-development} shows the development process to obtain the CDSL. First, a meta-model defining the CDSL entities and their relations is created using EMF. Then it is automatically translated to an Xtext grammar. Once the Xtext grammar is created, users can create their own CDSL models using the RoboComp DSL Editor, which performs code generation using M2T transformations. Because components need interfaces or topics  to communicate with other components, CDSL files can import data types, topics and interfaces defined in IDSL models (as shown in section~\ref{idsl}). 
\par
\begin{figure}[t!]
\begin{center}
\includegraphics[width=0.9\linewidth]{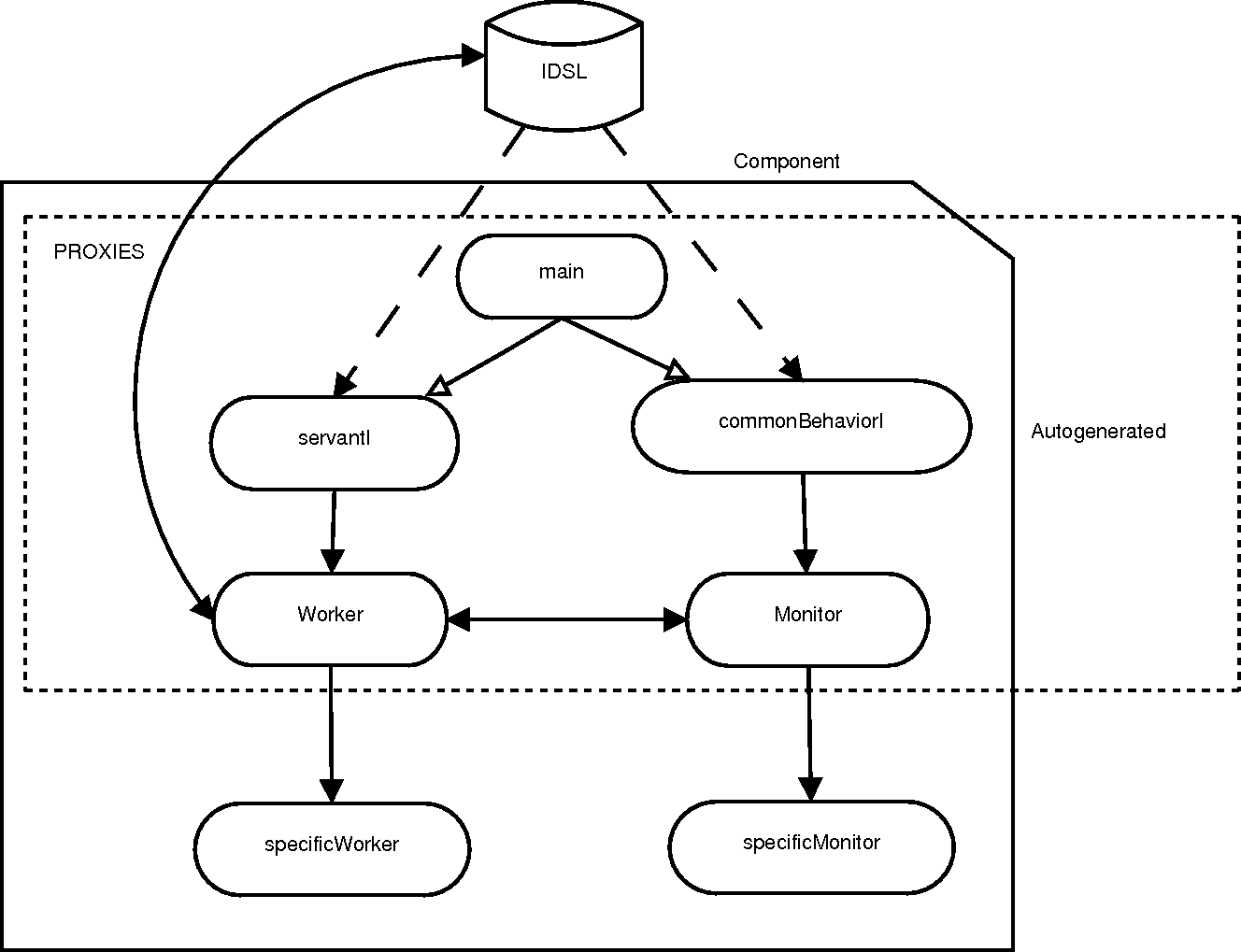}
\caption{Structure of RoboComp components.}
\label{fig:estructura_componente}
\end{center}
\end{figure}

The source code of the components generated from CDSL can be divided in two parts: the specific and the generic (see Figure~\ref{fig:estructura_componente}). The generic part contains the logic of interprocess communication, the general structure of the components (e.g., main program and threads, source directory structure, documentation rules or configuration parameters) and some introspection and self-monitoring capabilities. This generic functionality is implemented with abstract classes that are inherited and extended by the user-specific code to achieve the final working component. Thus, a component can be divided in two parts by a line separating the generic from the specific. The specific component code is generated by the RoboComp DSL Editor only the first time, but the generic code is always generated when regeneration from CDSL models occur. This way, users can be sure that their specific code will never be deleted and, at the same time, they are able to modify any component property. This organization is one of the most important design decisions. Figure~\ref{fig:study_cdsl} provides an example of the RoboComp DSL Editor while modifying a CDSL file.

\par
The properties that are contemplated in CDSL are the following:
\begin{itemize}
 \item Component name.
 \item Interfaces and data types defined in external IDSL files.
 \item Client/Server communication model: required and provided interfaces.
 \item Publish/Subscribe communication model: topics that the component will publish or subscribe to.
 \item Graphical interface support.
 \item State machine support.
 \item Dependences with external classes and libraries.
 \item Programming language of the component.
\end{itemize}

\subsection{IDSL}
\label{idsl}
RoboComp used the Ice Interface Definition Language (Slice by ZeroC) to define component interfaces. This language provides developers with a mechanism to create structured interfaces regardless of the platform and the programming language. Unfortunately, this language is middleware-dependent, so, if RoboComp had to be integrated with another middleware, all Slice files should be adapted. Moreover, any change in the Slice language would require modifying the RoboComp interfaces already defined. 
\par
To solve these problems, we developed the Interface Description Specific Language (IDSL). IDSL has been initially designed as a subset of the Slice features, mainly data types and procedures definition,  and avoiding interface inheritance, to facilitate future transformations among third party IDLs. Figure~\ref{dsl-development} shows the development process to obtain the IDSL, which is similar to CDSL. First the IDSL is defined using an EMF-based meta-model. Then the meta-model is imported to Xtext to generate the IDSL language. Users can define their own interfaces using the IDSL language, which is integrated in the RoboComp DSL Editor. Finally, they can generate the specific IDL using the code generator which is also integrated in the RoboComp DSL Editor. Figure~\ref{fig:study_idsl} provides a screenshot of the RoboComp DSL Editor while modifying an IDSL file.
\par
Currently, IDSL supports the following features:
\begin{itemize}
 \item Interfaces and topics definition.
 \item Basic data types, such as integer and real numbers or strings.
 \item Enumerated types.
 \item Custom structures and data types such as sequences and maps.
 \item Exceptions.
\end{itemize}

\subsection{DDSL}
\label{ddsl}
Components are independently executed programs that interact with each other. When using a component-oriented robotics framework, a robotic software system is composed of several interconnected components, representing a component network. These components can be executed manually, but as the number of components grows, it becomes increasingly difficult to manage them appropriately. Robots of middle complexity (e.g., mobile robots equipped with stereo heads)  and high complexity robots (e.g., mobile manipulators with expressive heads) are controlled by graphs containing dozens of components running on several computers. The configuration and management of these networks of processes suggest the combination of a graphical tool and a representation language.  The Deployment Description Specific Language was developed as the underlying language to make this management task easier.
\par
With DDSL, users are able to describe which components will be used, where they should be executed and which configuration to use. This makes it possible to automatically deploy RoboComp components in a certain computer or computer network. DDSL has been designed to simplify the system deployment and integration. Figure~\ref{dsl-development} shows the development process of the DDSL. It is defined using an EMF-based meta-model which is translated to an Xtext grammar. Once the Xtext DDSL grammar is created, developers can create and manage their DDSL models using the RoboComp DSL Editor. 
\par
In order to define a component network, the following parameters have to be specified for each component:
\begin{itemize}
 \item Component: the CDSL file path of the component to execute.
 \item Path to the executable file of the component.
 \item IP address and port.
 \item Path to the configuration file.
\end{itemize}
\par
With the information in this file, all component dependences can be precomputed. Thus, the DSL editor can warn users of basic configuration errors while editing the file and prior to the actual deployment.

\subsection{PDSL}
\label{pdsl}
The Parameter Definition Specific Language provides a generic structure for the configuration parameters that define the run-time behavior of the components. This DSL guides developers in writing the necessary configuration files in a standardized way within the framework. The file format that was previously used lacked of hierarchical structure, being just a list of {\textit{\textless{}attribute,value\textgreater{} pairs}}. When a component required a nested relation of parameters, such as a list of lists, the component had to parse the corresponding configuration string. Figure~\ref{fig:pdsl_listing} -particularly its last two lines- provides an example of this situation.
\begin{figure}[h!]
\begin{lstlisting}
# Endpoint
JointMotorComp.Endpoints=tcp -p 10067
# Parameters
JointMotor.NumMotors=2
JointMotor.Handler = Dunkermotoren
JointMotor.Device = /dev/ttyUSB0
JointMotor.BaudRate = 115200
JointMotor.BasicPeriod = 220
# Motor: name,id,invertedSign,min,max,zero,vel
JointMotor.M0 = dunker0,A,true,-3.14,3.14,0,.9
JointMotor.M1 = dunker1,B,true,-1.7,1.7,0,.9
\end{lstlisting}
\caption{Example of the structureless configuration file format used previously by RoboComp.}
\label{fig:pdsl_listing}
\end{figure}
\par
With PDSL, the configuration parameters can be organized in nested lists and the editor can check that the introduced values are within the predefined ranges. After the code generation process, the component can access the configuration information by using methods defined in the generic classes. This way, the programmer can safely access and modify the configuration values, according to their predefined value ranges. Figure~\ref{dsl-development} shows the development process of the PDSL. The dotted lines show that it is work in progress.

\begin{figure}[t!]
\begin{center}
\includegraphics[width=0.9\linewidth]{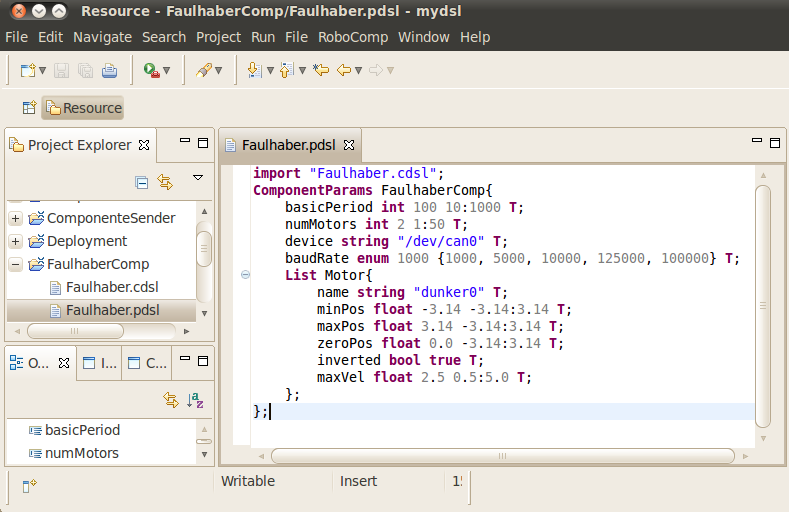}
\caption{Definition of the parameters using the PDSL}
\label{fig:pdsl-faulhaber}
\end{center}
\end{figure}
\par
Currently, PDSL supports the following parameter types:
\begin{itemize}
 \item Component: the CDSL file path of the component to execute.
 \item Basic data types, such as integer, booleans and real numbers or strings. 
 \item Enumerated types.
 \item List types and nested lists.
 \item Default values.
 \item Optional variables.
 \item Range of possible values   of basic and enumerated types.
\end{itemize}


\section{CASE STUDY}
\label{study}

\begin{figure}[t!]
\center
\includegraphics[width=0.49\textwidth]{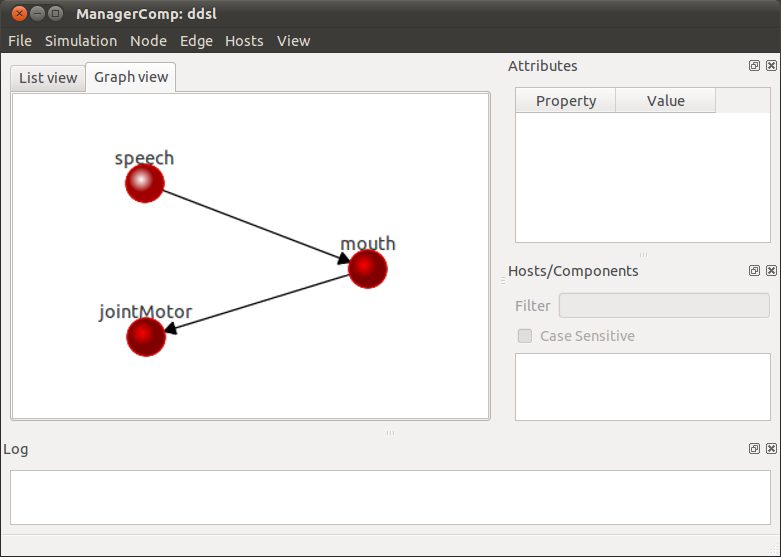}
\caption{Screenshot of the RCControlManager tool. The three components involved in the component network are shown. Once the user requests to run the Speech component, its dependencies are automatically satisfied.}
\label{fig:study_rcmanager}
\end{figure}

This section presents a real case study in which three common situations in component-oriented development are presented: the inclusion of an additional proxy in an existing component, the insertion of a new configuration parameter and the inclusion of a new method into an existing interface. These situations are framed in the following robotic context: starting with a speech synthesis component, the goal is to communicate it with another component that controls a robotic mouth. The final configuration should be able to synthesize text and to move the mouth synchronously. Figure~\ref{fig:study_rcmanager} shows all the components involved as seen from the RCControlManager deployment utility. As introduced in section~\ref{robocomp}, before making RoboComp DSL-based, to complete these changes it was required to make several changes in the source code manually. In this section we will show how the use of DSLs reduces the time to introduce these changes, avoids unwanted errors in the code and improves the user experience. The steps required to perform all the modifications are:

\begin{itemize}
 \item Modify the Component Description file to include a new proxy to MouthComp.
 \item Include a new parameter in the Parameters Definition file of SpeechComp to specify if the connection to MouthComp will be used.
 \item Add a new entry in the Deployment Definition file to include the component MouthComp.
 \item Optionally, add a new method in the interface of SpeechComp to activate/deactivate in run-time the mouth synchronization.
\end{itemize}

\par

\begin{figure}[t!]
\center
\includegraphics[width=0.49\textwidth]{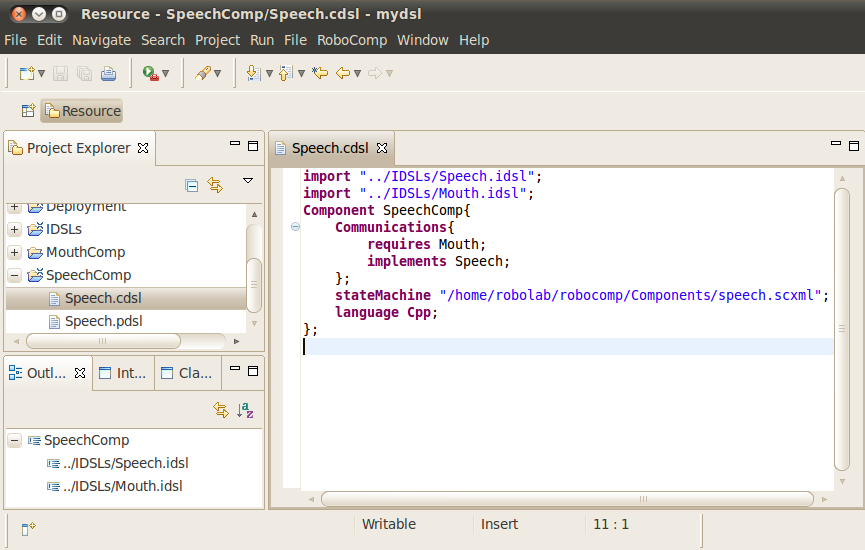}
\caption{Screenshot of the RoboComp DSL Editor while modifying the CDSL model in the case study.}
\label{fig:study_cdsl}
\end{figure}

The first step to introduce an additional proxy to the Speech component is to change its CDSL file (i.e., the file describing the generic properties of the component). This can be done using the RoboComp DSL Editor (see Figure~\ref{fig:study_cdsl}). After re-generating the generic code, the specific classes will automatically have access to the new proxy (no changes in the specific classes are needed for it).
\par

\par
\begin{figure}[t!]
\center
\includegraphics[width=0.49\textwidth]{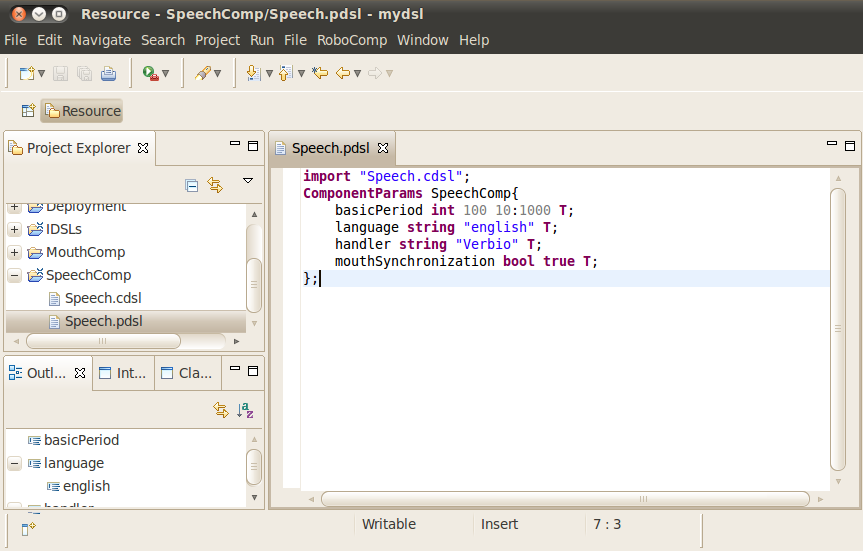}
\caption{Screenshot of the RoboComp DSL Editor while modifying the PDSL model in the case study.}
\label{fig:study_pdsl}
\end{figure}

After modifying the CDSL file, and only if the connection to the Mouth component is going to be optional, the PDSL file must be updated to include the corresponding variable. In this case, we created a new boolean parameter named mouthSynchronization as shown in Figure~\ref{fig:study_pdsl}.

\par

Once the code has been re-generated and the SpeechComp configuration parameters have been configured, the component is ready to work. However, since it depends on the execution of MouthComp, this component must be previously executed. Deployment can be done manually, but it is a hard task when the component network is composed of more than a few components. The RoboComp DSL Editor can be used to specify different deployment scenarios. In this case the network is composed by only three components: SpeechComp, MouthComp and JointMotorComp (which is a dependence of the MouthComp component). RCControlManager, the component manager of RoboComp, takes the DDSL file as input and manages the execution of the components and the dependencies among them.

\begin{figure}[t!]
\center
\includegraphics[width=0.49\textwidth]{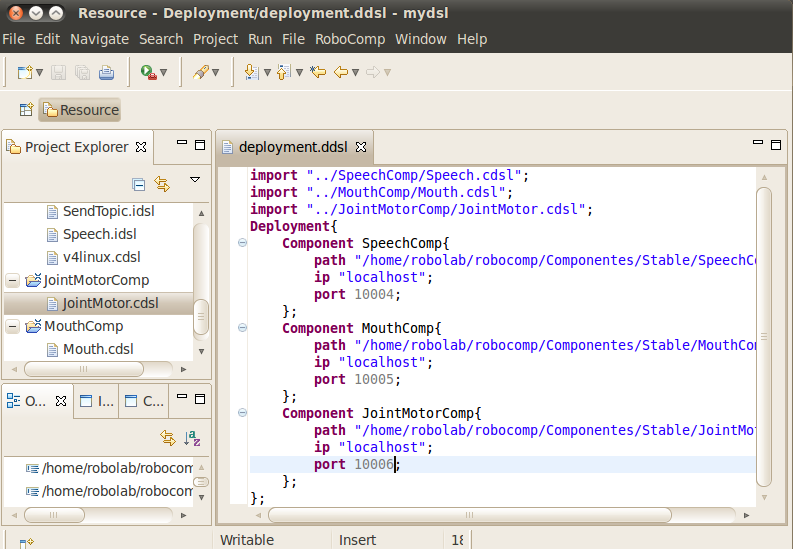}
\caption{Screenshot of the RoboComp DSL Editor while modifying the DDSL model in the case study.}
\label{fig:study_ddsl}
\end{figure}

Figures~\ref{fig:study_ddsl} and~\ref{fig:study_rcmanager} provide screenshots of the RoboComp DSL Editor while modifying the previous DDSL file and the RCControlManager tool, respectively.

\par
RCControlManager reads the deployment file and generates a visual graph. Components are shown as nodes and component dependencies as edges. Moreover, the deployment file can be generated or modified graphically using this tool. When a new host is added, RCControlManager checks for the availability of any known component in that machine and shows them in a list. A deployment configuration is visually created dropping components from this list. Dependences may be assigned by dragging nodes over target nodes. Deployment files are created very quickly using this technology.

\begin{figure}[t!]
\center
\includegraphics[width=0.49\textwidth]{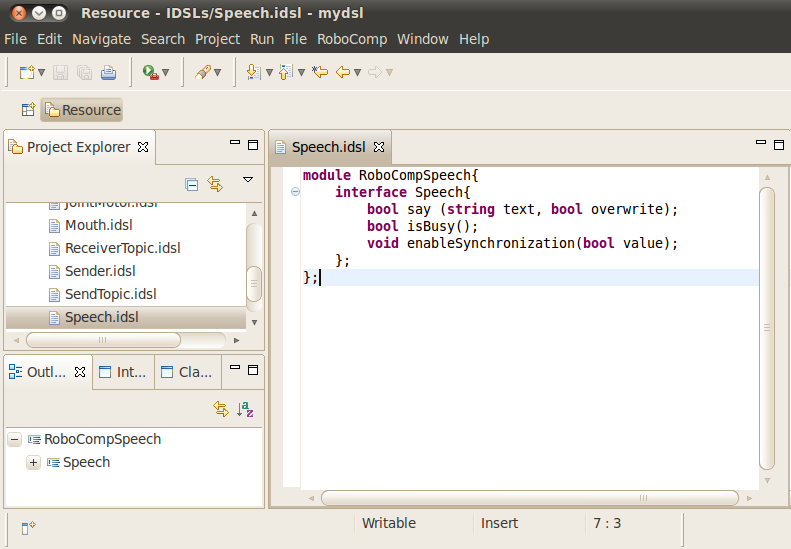}
\caption{Screenshot of the RoboComp DSL Editor while modifying the IDSL model in the case study.}
\label{fig:study_idsl}
\end{figure}

Even though it may not be actually necessary, the developer might want to include a new method in the SpeechComp interface in order to activate/deactivate the synchronization with the robotic mouth on-line. In this case, there are three steps to take: {\bf a)} use the RoboComp DSL Editor to include the line corresponding to the new method (see Figure~\ref{fig:study_idsl}); {\bf b)} re-generate the IDSL file in order to obtain the new IDL implementation; {\bf c)} re-generate the SpeechComp CDSL to include the new function in the generic code. It is very important to note that this is the only step in which users need to modify the specific code manually.

\section{EXPERIMENTAL RESULTS}
\label{experiments}
An experiment was conducted in order to provide empirical evidences supporting the claims made in the paper. Twelve roboticists who were already familiar with the framework were asked to perform five of the most common repetitive tasks they have to face when developing and managing robotics components. In particular, they were asked to make the following changes and operations with an existing robotics component: {\bf a)} to create a new proxy for a given interface; {\bf b)} to make the component provide a new additional interface; {\bf c)} to include a new method in the previous interface; {\bf d)} to include a new library and a new class in the component project; and {\bf e)} to deploy a small component network.
\par
In order to be able to evaluate the benefits of the proposed approach, the experiments were performed twice: first, using the DSL technology and then, without it. For all the experiments two variables were measured: the time spent performing the task and the lines of code written. In the case of the last task (deploying a component network), the number of commands executed where measured instead of the lines of code written.
\par
\begin{figure}[t!]
\center
\includegraphics[width=0.49\textwidth]{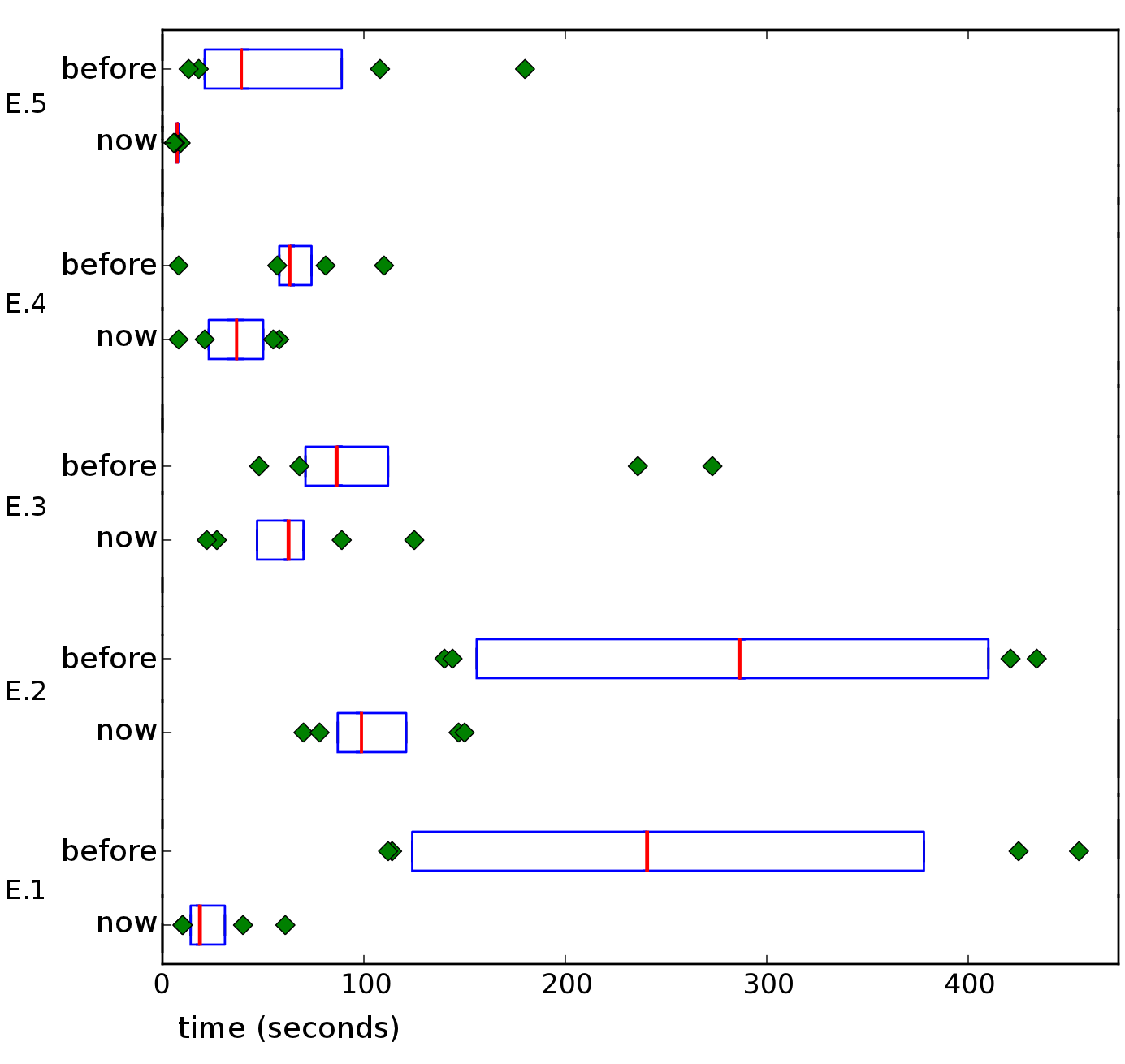}
\caption{Graph showing the corresponding boxplots for the times associated with the five experiments. For each experiment E.\textit{n}, it is shown the time spent with and without using the DSL (before and now, respectively). }
\label{fig:experiments_time}
\end{figure}
\begin{figure}[t!]
\center
\includegraphics[width=0.49\textwidth]{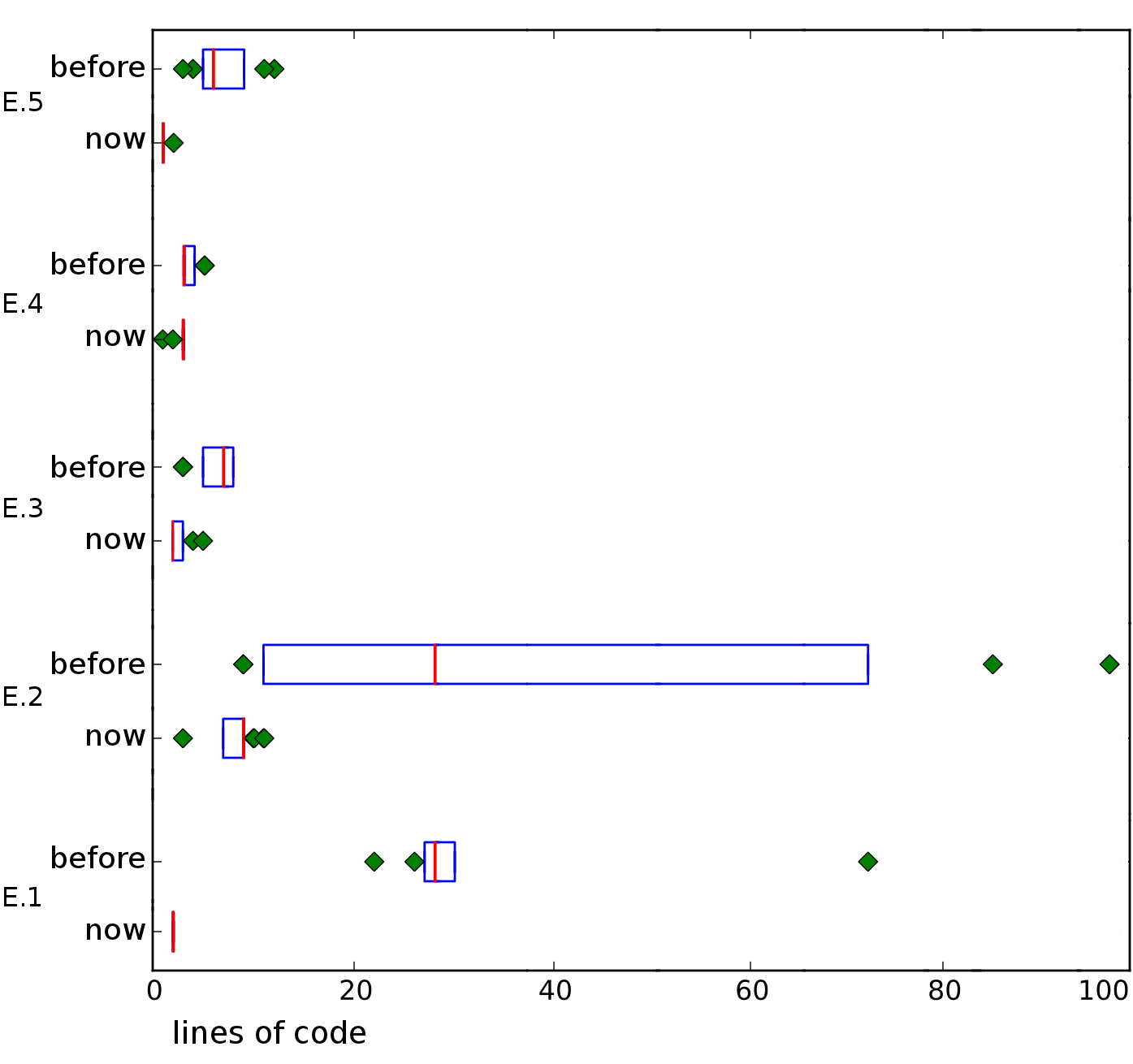}
\caption{Graph showing the corresponding boxplots for the lines of code associated with the five experiments. For each experiment E.\textit{n} it is shown the lines of code written with and without using the DSL (before and now, respectively).}
\label{fig:experiments_loc}
\end{figure}

The data obtained from the time employed and the lines of code written is displayed in figures~\ref{fig:experiments_time} and~\ref{fig:experiments_loc}, respectively. Measurement data is represented as boxplots containing all measurements ranging between the first and third quartiles. The location of the median of the measurements is indicated by a red line crossing the rectangle vertically. Measurements outside the box are considered outliers and are drawn using green diamonds. 
\par
Figure~\ref{fig:experiments_time} shows the results regarding the time spent in the experiments. It is worth mentioning that the only time taken into account was the one in which the subject was typing code, not thinking. Since users need less time to think when using DSLs (i.e., there is no need to think which code pieces should they change) this plays against the use of DSLs. Inspite of this, it can be seen how using the DSL approach shorter times were achieved for all of the experiments performed.
\par
Figure~\ref{fig:experiments_loc} shows the results regarding the lines of code written while performing the experiments. As happened with time, the figure shows that, using the DSL approach, fewer lines of code were written for all of the experiments performed. This is not a surprise, since small changes in the DSLs might involve many changes in the generated code. The only experiment in which no considerable improvements were achieved (only a few seconds) was the experiment number 3. This is because the framework was already making use of CMake features for the operations involved in the experiment, so the initial number of lines to modify was already low.
\par

\section{CONCLUSIONS AND FUTURE WORKS}
\label{conclusions}
\subsection{Conclusions}
This paper has presented four DSLs based on the MDA approach used to improve the lifecycle of robotics components: {\bf a)} for specifying general component properties (CDSL); {\bf b)} for defining component interfaces (IDSL); {\bf c)} for the deployment of components (DDSL); and {\bf d)} to define and initialize component parameters (PDSL).
\par
CDSL allows components to be created and modified even into different languages. It reduces the workload, so developers can center their effort in the implementation of the behaviour of the components they are developing. IDSL is a first step to make RoboComp independent not only of the architecture and the programming language, but also of the middleware. DDSL eases the deployment stage of the lifecycle of the components. It describes a network of components and their dependencies, as well as where and how should they be executed. This allows RCControlManager, the component manager of RoboComp, to run a component network with just a mouse click. Finally, PDSL provides a means to store the configuration parameters of the components in a structured and less error-prone way.
\par
These DSLs have been used on the RoboComp framework to improve its flexibility, scalability and maintenance, making it possible to create and manage components in a higher level of abstraction. Moreover, developers can benefit from M2M and M2T in order to perform transformations between models and obtain the final source code automatically. These DSLs have been developed as textual models in order to reduce the development time, although they can be used in conjunction with visual models.
\par
All the software described in this paper is freely available for download from the RoboComp site: http://robocomp.sf.net/DSLRob2011.
\subsection{Future Works}
One of the limitations of many current robotics frameworks is that they are middleware dependent. Thus, components can only communicate with those built using the same framework or, in some cases, those using the same middleware. Although there are some efforts towards building bridges between frameworks, it is not clear how these bridges will provide all the necessary functionality. To overcome this limitation, efforts are currently being made towards achieving middleware-independence and rely, instead, on different reference implementations.
\par
Despite the powerful component and configuration models provided by RoboComp, its wide adoption could be compromised if it does not fit well the stringent real-time and performance requirements of a distributed real-time embedded system. In this sense, transparent support for DDS~\cite{DDS} (OMG's Data Distribution Service for Real-Time systems) is underway and initial encouraging results have already been published~\cite{simpar}. This standard addresses the need for real-time and quality of service features in distributed applications and is being adopted in mission and business-critical applications, such as air traffic control, telemetry or financial trading systems.
\par
Another improvement we are working on is the hierarchical representation of groups of components. In ongoing work, one component is automatically created to act as a proxy for all  incoming communications to the group. This rearrangement makes a group of components appear as a single one to the rest, at the cost of a certain increase in delay. We believe this additional level of abstraction is necessary to handle dozens of running components, a common situation in future complex robotic scenarios.   
\par


\end{document}